\documentclass{article}

\usepackage{microtype}
\usepackage{graphicx}
\usepackage{subfigure}
\usepackage{booktabs} % for professional tables
\usepackage{multirow}

\usepackage{hyperref}

\usepackage[nolist]{acronym}
\begin{acronym}[DRAM]
    \acro{AI}{artificial intelligence}
    \acrodefindefinite{AI}{an}{an}
    \acro{ASIC}{application-specific integrated circuit}
    \acrodefindefinite{ASIC}{an}{an}
    \acro{BD}{bounded delay}
    \acro{BNN}{binarized neural network}
    \acro{CNN}{convolutional neural network}
    \acro{CTM}{convolutional Tsetlin machine}
    \acro{FSM}{finite state machine}
    \acro{LA}{learning automaton}
    \acrodefplural{LA}{learning automata}
    \acrodefindefinite{LA}{an}{a}
    \acro{ML}{machine learning}
    \acrodefindefinite{ML}{an}{a}
    \acro{TA}{Tsetlin automaton}

    \acrodefplural{TA}{Tsetlin automata}
    \acro{TAT}{Tsetlin automaton team}
    \acro{TM}{Tsetlin machine}
    \acro{RTM}{regression Tsetlin machine}
\end{acronym}

\usepackage[accepted]{icml2021}

\icmltitlerunning{Massively Parallel and Asynchronous Tsetlin Machine Architecture}

\begin{document}

\twocolumn[
\icmltitle{Massively Parallel and Asynchronous Tsetlin Machine Architecture Supporting Almost Constant-Time Scaling}



\icmlsetsymbol{equal}{*}

\begin{icmlauthorlist}
\icmlauthor{Kuruge Darshana Abeyrathna}{equal,to}
\icmlauthor{Bimal Bhattarai}{equal,to}
\icmlauthor{Morten Goodwin}{equal,to}
\icmlauthor{Saeed Rahimi Gorji}{equal,to}
\icmlauthor{Ole-Christoffer Granmo}{equal,to}
\icmlauthor{Lei Jiao}{equal,to}
\icmlauthor{Rupsa Saha}{equal,to}
\icmlauthor{Rohan Kumar Yadav}{equal,to}
\end{icmlauthorlist}

\icmlaffiliation{to}{Department of Information and Communication Technology, Unviersity of Agder, Grimstad, Norway}

\icmlcorrespondingauthor{Ole-Christoffer Granmo}{ole.granmo@uia.no}

\icmlkeywords{Machine Learning, ICML}

\vskip 0.3in
]



\printAffiliationsAndNotice{\icmlEqualContribution({The authors are ordered alphabetically by last name.})}

\begin{abstract}
Using logical clauses to represent patterns, \acp{TM} have recently obtained competitive performance in terms of accuracy, memory footprint, energy, and learning speed on several benchmarks. Each \ac{TM} clause votes for or against a particular class, with classification resolved using a majority vote. While the evaluation of clauses is fast, being based on binary operators, the voting makes it necessary to synchronize the clause evaluation, impeding parallelization. In this paper, we propose a novel scheme for desynchronizing the evaluation of clauses, eliminating the voting bottleneck. In brief, every clause runs in its own thread for massive native parallelism. For each training example, we keep track of the class votes obtained from the clauses in local voting tallies. The local voting tallies allow us to detach the processing of each clause from the rest of the clauses, supporting decentralized learning. This means that the \ac{TM} most of the time will operate on outdated voting tallies. We evaluated the proposed parallelization across diverse learning tasks and it turns out that our decentralized \ac{TM} learning algorithm copes well with working on outdated data, resulting in no significant loss in learning accuracy. Furthermore, we show that the proposed approach provides up to $50$ times faster learning. Finally, learning time is almost constant for reasonable clause amounts (employing from $20$ to $7,000$ clauses on a Tesla V100 GPU). For sufficiently large clause numbers, computation time increases approximately proportionally. Our parallel and asynchronous architecture thus allows processing of massive datasets and operating with more clauses for higher accuracy.
\end{abstract}

\section{Introduction}
Tsetlin machines~(\acp{TM})~ \cite{granmo2018tsetlin} have recently demonstrated competitive results in terms of accuracy, memory footprint, energy, and learning speed on diverse benchmarks (image classification, regression, natural language understanding, and speech processing) \cite{berge2019text,rohan2021AAAI,abeyrathna2020nonlinear,granmo2019convtsetlin,wheeldon2020learning,abeyrathna2021integer,lei2021kws}.
They use frequent pattern mining and resource allocation principles to extract common patterns in the data, rather than relying on minimizing output error, which is prone to overfitting.  Unlike the intertwined nature of pattern representation in neural networks, a \ac{TM} decomposes problems into self-contained patterns, expressed as conjunctive clauses in propositional logic (i.e., in the form \textbf{if} input $X$ \textbf{satisfies} condition $A$ \textbf{and not} condition $B$ \textbf{then} output $y = 1$). The clause outputs, in turn, are combined into a classification decision through summation and thresholding, akin to a logistic regression function, however, with binary weights and a unit step output function. Being based on the human-interpretable disjunctive normal form \cite{valiant1984learnable}, like Karnaugh maps \cite{karnaugh1953map}, a \ac{TM} can map an exponential number of input feature value combinations to an appropriate output \cite{granmo2018tsetlin}.

\paragraph{Recent progress on \acp{TM}} Recent research reports several distinct TM properties. The TM can be used in convolution, providing competitive performance on MNIST, Fashion-MNIST, and Kuzushiji-MNIST, in comparison with CNNs, K-Nearest Neighbor, Support Vector Machines, Random Forests, Gradient Boosting, BinaryConnect, Logistic Circuits and ResNet \cite{granmo2019convtsetlin}. The \ac{TM} has also achieved promising results in text classification \cite{berge2019text}, word sense disambiguation \cite{yadav2021wordsense}, novelty detection \cite{bhattarai2021novelty, bhattarai2021wordlevel}, fake news detection \cite{bhattarai2021explainable}, semantic relation analysis \cite{saha2020causal},  and aspect-based sentiment analysis \cite{rohan2021AAAI} using the conjunctive clauses to capture textual patterns. Recently, regression \acp{TM} compared favorably with Regression Trees, Random Forest Regression, and Support Vector Regression \cite{abeyrathna2020nonlinear}. The above \ac{TM} approaches have further been enhanced by various techniques. By introducing real-valued clause weights, it turns out that the number of clauses can be reduced by up to $50\times$ without loss of accuracy \cite{phoulady2020weighted}. Also, the logical inference structure of \acp{TM} makes it possible to index the clauses on the features that falsify them, increasing inference- and learning speed by up to an order of magnitude \cite{gorji2020indexing}. Multi-granular clauses simplify the hyper-parameter search by eliminating the pattern specificity parameter \cite{gorji2019multigranular}.  In \cite{abeyrathna2021integer}, stochastic searching on the line automata \cite{oommen1997stochastic} learn integer clause weights, performing on-par or better than Random Forest, Gradient Boosting, Neural Additive Models, StructureBoost and Explainable Boosting Machines.  Closed form formulas for both local and global \ac{TM} interpretation, akin to SHAP, was proposed by \citet{blakely2020closedform}. From a hardware perspective, energy usage can be traded off against accuracy by making inference deterministic \cite{abeyrathna2020deterministic}. Additionally, \citet{shafik2020explainability} show that \acp{TM} can be fault-tolerant, completely masking stuck-at faults. Recent theoretical work proves convergence to the correct operator for ``identity" and ``not". It is further shown that arbitrarily rare patterns can be recognized, using a quasi-stationary Markov chain-based analysis. The work finally proves that when two patterns are incompatible, the most accurate pattern is selected \cite{zhang2020convergence}. Convergence for the ``XOR" operator has also recently been proven by~\citet{jiao2021convergence}.

\paragraph{Paper Contributions}
In all of the above mentioned \ac{TM} schemes, the clauses are learnt using \ac{TA}-teams~\cite{Tsetlin1961} that interact to build and integrate conjunctive clauses for decision-making. While producing accurate learning, this interaction creates a bottleneck that hinders parallelization. That is, the clauses must be evaluated and compared before feedback can be provided to the \acp{TA}.

In this paper, we first cover the basics of \acp{TM} in Section~\ref{sec:basics}. Then, we propose a novel parallel and asynchronous architecture in Section~\ref{sec:architecture}, where every clause runs in its own thread for massive parallelism. We eliminate the above interaction bottleneck by introducing local voting tallies that keep track of the clause outputs, per training example. The local voting tallies detach the processing of each clause from the rest of the clauses, supporting decentralized learning. Thus, rather than processing training examples one-by-one as in the original \ac{TM}, the clauses access the training examples simultaneously, updating themselves and the local voting tallies in parallel. In Section~\ref{sec:empirical}, we investigate the properties of the new architecture empirically on regression, novelty detection,
%reinforcement learning,
semantic relation analysis and word sense disambiguation. We show that our decentralized \ac{TM} architecture copes well with working on outdated data, with no measurable loss in learning accuracy. We further investigate how processing time scales with the number of clauses, uncovering almost constant-time processing over reasonable clause amounts. Finally, in Section~\ref{sec:conclusion}, we conclude with pointers to future work, including architectures for grid-computing and heterogeneous systems spanning the cloud and the edge.

The main contributions of the proposed architecture can be summarized as follows: 
\begin{itemize}
\item Learning time is made almost \emph{constant} for reasonable clause amounts (employing from $20$ to $7,000$ clauses on a Tesla V100 GPU).
\item For sufficiently large clause numbers, computation time increases approximately proportionally
%%From feedback - Reviewer 5 Q1
to the increase in number of clauses.
%%From feedback - Reviewer 5 Q1
\item The architecture copes remarkably with working on outdated data, resulting in no significant loss in learning accuracy across diverse learning tasks (regression, novelty detection,
%reinforcement learning,
semantic relation analysis, and word sense disambiguation).
\end{itemize}
Our parallel and asynchronous architecture thus allows processing of more massive data sets and operating with more clauses for higher accuracy, significantly increasing the impact of logic-based machine learning.
\section{Tsetlin Machine Basics}\label{sec:basics}

\subsection{Classification}

A \ac{TM} takes a vector $X=[x_1,\ldots,x_o]$ of $o$ Boolean features as input, to be classified into one of two classes, $y=0$ or $y=1$. These features are then converted into a set of literals that consists of the features themselves as well as their negated counterparts: $L = \{x_1,\ldots,x_o,\neg{x}_1,\ldots,\neg{x}_o\}$.

If there are $m$ classes and $n$ sub-patterns per class, a TM employs \(m \times n\) conjunctive clauses to represent the sub-patterns. For a given class\footnote{Without loss of generality, we consider only one of the classes, thus simplifying notation. Any \ac{TM} class is modelled and processed in the same way.}, we index its clauses by $j$, \(1 \leq j \leq n\), each clause being a conjunction of literals:
\begin{equation}
\textstyle
C_j (X)=\bigwedge_{l_k \in L_j} l_k.
\end{equation}
\noindent Here, $l_k, 1 \leq k \leq 2o,$ is a feature or its negation. Further,  \(L_j\) is a subset of the literal set \(L\). For example, the particular clause $C_j(X) = x_1 \land \lnot x_2$ consists of the literals $L_j = \{x_1, \lnot x_2\}$ and outputs $1$ if $x_1 = 1$ and $x_2 = 0$.

The number of clauses $n$ assigned to each class is user-configurable. The clauses with odd indexes are assigned positive polarity and the clauses with even indexes are assigned negative polarity. The clause outputs are combined into a classification decision through summation and thresholding using the unit step function $u(v) = 1 ~\mathbf{if}~ v \ge 0 ~\mathbf{else}~ 0$:
\begin{equation}\label{eqn:voting}
\textstyle
\hat{y}=u\left(\Sigma_{j=1,3,\ldots}^{n-1}C_j{(X)}-\Sigma_{i=2,4,\ldots}^{n}C_j{(X)}\right).
\end{equation}

Namely, classification is performed based on a majority vote, with the positive clauses voting for $y=1$ and the negative for $y=0$. 
%The classifier $\hat{y} = u\left(x_1 \bar{x}_2 + \bar{x}_1 x_2 - x_1 x_2 - \bar{x}_1 \bar{x}_2\right)$, e.g., captures the XOR-relation.

\begin{figure}[ht]
\vskip 0.2in
\begin{center}
\centerline{\includegraphics[width=\columnwidth]{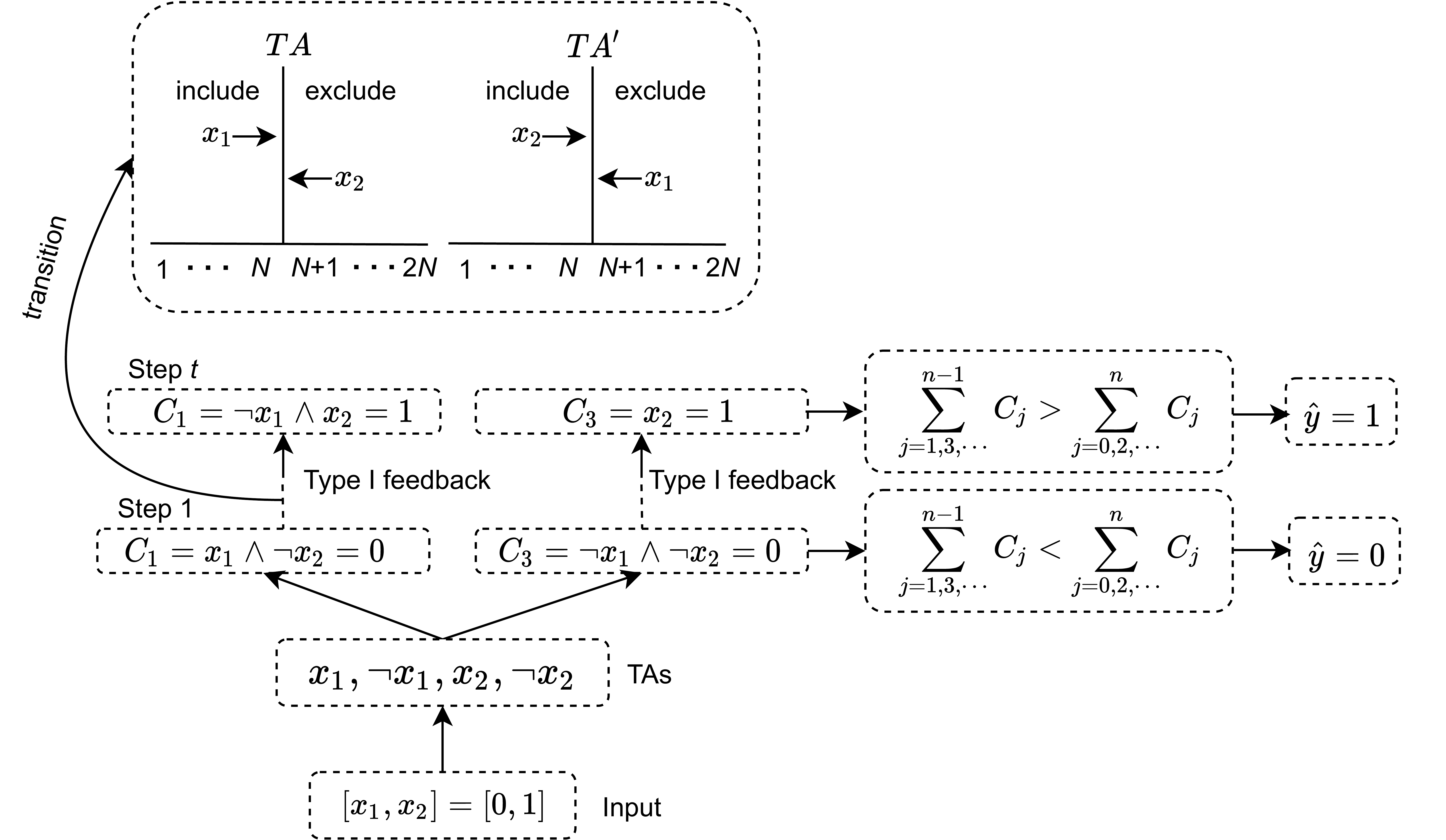}}
\caption{\ac{TM} learning dynamics for an XOR-gate training sample, with input ($x_1=0, x_2=1$) and output target $y=1$.}
\label{figure:tm_architecture_basic}
\end{center}
\vskip -0.2in
\end{figure}

\subsection{Learning}

\ac{TM} learning is illustrated in Fig. \ref{figure:tm_architecture_basic}. As shown, a clause $C_j(X)$ is composed by a team of \acp{TA}. Each \ac{TA} has $2N$ states and decides to \emph{Include} (from state $1$ to $N$) or \emph{Exclude} (from state $N+1$ to $2N$ ) a specific literal $l_k$ in the clause. In the figure, \ac{TA} refers to the \acp{TA} that control the original form of a feature ($x_1$ and $x_2$) while \ac{TA}' refers to those controlling negated features ($\lnot x_1$ and $\lnot x_2$). A \ac{TA} updates its state based on the feedback it receives in the form of Reward, Inaction, and Penalty (illustrated by the features moving in a given direction in the TA-part of the figure). There are two types of feedback associated with \ac{TM} learning: Type~I feedback and Type~II feedback,  which are shown in Table~\ref{table1} and Table \ref{table22}, respectively. \par
\textbf{Type I feedback} is given stochastically to clauses with odd indexes when $y=1$ and to clauses with even indexes when $y=0$. Each clause, in turn, reinforces its \acp{TA} based on: (1) its output $C_j(X)$; (2) the action of the \ac{TA} -- \emph{Include} or \emph{Exclude}; and (3) the value of the literal $l_k$ assigned to the \ac{TA}. As shown in Table~\ref{table1}, two rules govern Type I feedback:
\begin{itemize}
\item \emph{Include} is rewarded and \emph{Exclude} is penalized with probability $\frac{s-1}{s}~\mathbf{if}~C_j(X)=1~\mathbf{and}~l_k=1$. This reinforcement is strong\footnote{Note that the probability $\frac{s-1}{s}$ is replaced by $1$ when boosting true positives.} (triggered with high probability) and makes the clause remember and refine the pattern it recognizes in $X$. 
\item \emph{Include} is penalized and \emph{Exclude} is rewarded with probability $\frac{1}{s}~\mathbf{if}~C_j(X)=0~\mathbf{or}~l_k=0$. This reinforcement is weak (triggered with low probability) and coarsens infrequent patterns, making them frequent.
\end{itemize}
Above, parameter $s$ controls pattern frequency.

\textbf{Type II feedback} is given stochastically to clauses with odd indexes when $y\!=\!0$ and to clauses with even indexes when $y\!=\!1$. As captured by Table~\ref{table22}, it penalizes \emph{Exclude} with probability $1$ $\mathbf{if}~C_j(X)=1~\mathbf{and}~l_k=0$. Thus, this feedback produces literals for discriminating between $y=0$ and $y=1$.

\begin{table}[t]
\centering
\vskip 0.15in
\begin{center}
\begin{small}
\begin{sc}
\resizebox{.95\columnwidth}{!}{
\begin{tabular}{l|l|l|l}
    \hline
    \multirow{2}{*}{Input}&Clause & \ \ \ \ \ \ \ 1 & \ \ \ \ \ \ \ 0 \\
    &{Literal} &\ \ 1 \ \ \ \ \ \ 0 &\ \ 1 \ \ \ \ \ \ 0 \\
    \hline
    \multirow{2}{*}{Include Literal}&P(Reward)&$\frac{s-1}{s}$\ \ \ NA & \ \ 0 \ \ \ \ \ \ 0\\ [1mm]
    &P(Inaction)&$\ \ \frac{1}{s}$\ \ \ \ \ NA &$\frac{s-1}{s}$ \ $\frac{s-1}{s}$ \\ [1mm]
    &P(Penalty)& \ \ 0 \ \ \ \ \ NA& $\ \ \frac{1}{s}$ \ \ \ \ \  $\frac{1}{s}$ \\ [1mm]
    \hline
    \multirow{2}{*}{Exclude Literal}&P(Reward)& \ \ 0 \ \ \ \ \ \ $\frac{1}{s}$ & $\ \ \frac{1}{s}$ \ \ \ \ \
    $\frac{1}{s}$ \\ [1mm]
    &P(Inaction)&$ \ \ \frac{1}{s}$\ \ \ \ $\frac{s-1}{s}$  &$\frac{s-1}{s}$ \ $\frac{s-1}{s}$ \\ [1mm]
    &P(Penalty)&$\frac{s-1}{s}$ \ \ \ \ 0& \ \ 0 \ \ \ \ \ \ 0 \\ [1mm]
    \hline
\end{tabular}
}
\end{sc}
\end{small}
\end{center}
\vskip -0.1in
\caption{Type I Feedback}
\label{table1}
\end{table}

\begin{table}[t]
\centering
\vskip 0.15in
\begin{center}
\begin{small}
\begin{sc}
\resizebox{.95\columnwidth}{!}{
\begin{tabular}{l|l|l|l}
    \hline
    \multirow{2}{*}{Input}&Clause & \ \ \ \ \ \ \ 1 & \ \ \ \ \ \ \ 0 \\
    &{Literal} &\ \ 1 \ \ \ \ \ \ 0 &\ \ 1 \ \ \ \ \ \ 0 \\
    \hline
    \multirow{2}{*}{Include Literal}&P(Reward)&\ \ 0 \ \ \ NA & \ \ 0 \ \ \ \ \ \ 0\\[1mm]
    &P(Inaction)&1.0 \ \  NA &  1.0 \ \ \ 1.0 \\[1mm]
    &P(Penalty)&\ \ 0 \ \ \ NA & \ \ 0 \ \ \ \ \ \ 0\\[1mm]
    \hline
    \multirow{2}{*}{Exclude Literal}&P(Reward)&\ \ 0 \ \ \ \ 0 & \ \ 0 \ \ \ \ \ \ 0\\[1mm]
    &P(Inaction)&1.0 \ \ \ 0 &  1.0 \ \ \ 1.0 \\[1mm]
    &P(Penalty)&\ \ 0 \ \  1.0 & \ \ 0 \ \ \ \ \ \ 0\\[1mm]
    \hline
\end{tabular}
}
\end{sc}
\end{small}
\end{center}
\vskip -0.1in
\caption{Type II Feedback}
\label{table22}
\end{table}

%%From feedback - {Reviewer 8. Comment regarding how rewards/penalties are actualized.} 
\par The ``state" is realized as a simple counter per Tsetlin Automaton. In practice, reinforcing Include (penalizing Exclude or rewarding Include) is done by increasing the counter, while reinforcing exclude (penalizing Include or rewarding Exclude) is performed by decreasing the counter.
%%From feedback - {Reviewer 8. Comment regarding how rewards/penalties are actualized.} 
\par As an example of learning, let us consider a dataset with XOR-gate sub-patterns. In particular, consider the input \((x_1 = 0, x_2 = 1)\) and target output \(y= 1\) to visualize the learning process (Fig.~\ref{figure:tm_architecture_basic}). We further assume that we have $n=4$ clauses per class. Among the $4$ clauses, the clauses $C_1$ and $C_3$ vote for $y=1$ and the clauses $C_0$ and $C_2$ vote for $y=0$. For clarity, let us only consider how $C_1$ and $C_3$ learn  a sub-pattern from the given sample of XOR-gate input and output. At Step~1 in the figure, the clauses have not yet learnt the pattern for the given sample. This leads to the wrong class prediction \((\hat{y}=0)\), thereby triggering Type~I feedback for the corresponding literals. Looking up clause output $C_1(X) = 0$ and literal value $x_1=0$ in Table~\ref{table1}, we note that the \ac{TA} controlling $x_1$ receives either Inaction or Penalty feedback for including $x_1$ in $C_1$, with probability $\frac{s-1}{s}$ and $\frac{1}{s}$, respectively. After receiving several penalties, with high probability, the  \ac{TA} changes its state to selecting \emph{Exclude}. Accordingly, literal $x_1$ gets removed from the clause $C_1$. After that, the \ac{TA} that has excluded literal $\neg{x_1}$ from $C_1$ also obtains penalties, and eventually switches to the \emph{Include} side of its state space. The combined outcome of these updates are shown in Step $t$ for $C_1$. Similarly, the \ac{TA} that has included literal $\neg{x_2}$ in clause $C_1$ receives Inaction or Penalty feedback with probability $\frac{s-1}{s}$ and $\frac{1}{s}$, respectively. After obtaining multiple penalties, with high probability, $\neg{x_2}$ becomes excluded from $C_1$. Simultaneously, the \ac{TA} that controls $x_2$ ends up in the \emph{Include} state, as also shown in Step $t$. At this point,  both clause $C_1$ and $C_3$ outputs $1$ for the given input, correctly predicting the output $\hat{y}=1$. 

\textbf{Resource allocation} dynamics ensure that clauses distribute themselves across the frequent patterns, rather than missing some and over-concentrating on others. That is, for any input $X$, the probability of reinforcing a clause gradually drops to zero as the clause output sum
\begin{equation}\label{eqn:aggregated_votes}
\textstyle
    v = \Sigma_{j=1,3,\ldots}^{n-1}C_j{(X)}-\Sigma_{i=2,4,\ldots}^{n}C_j{(X)}
\end{equation}
approaches a user-set margin $T$ for $y=1$ (and $-T$ for $y=0$). 
If a clause is not reinforced, it does not give feedback to its \acp{TA}, and these are thus left unchanged.  In the extreme, when the voting sum $v$ equals or exceeds the margin $T$ (the TM has successfully recognized the input $X$), no clauses are reinforced. They are then free to learn new patterns, naturally balancing the pattern representation resources  \cite{granmo2018tsetlin}.
\section{Parallel and Asynchronous Architecture}\label{sec:architecture}
Even though CPUs have been traditionally geared to handle high workloads, they are more suited for sequential processing and their performance is still dependant on the limited number of cores available. In contrast, since GPUs are primarily designed for graphical applications by employing many small processing elements, they offer a large degree of parallelism \cite{owens2007survey}. As a result, a growing body of research has been focused on performing general purpose GPU computation or GPGPU. For efficient use of GPU power, it is critical for the algorithm to expose a large amount of fine-grained parallelism \cite{jiang2005automatic,satish2009designing}.

While the voting step of the \ac{TM} (Eq. \ref{eqn:aggregated_votes}) hinders parallelization, the remainder of the \ac{TM} architecture is natively parallel.  In this section, we introduce our decentralized inference scheme and the accompanying architecture that makes it possible to have parallel asynchronous learning and classification, resolving the voting bottleneck by using local voting tallies.

\subsection{Voting Tally}

A voting tally that tracks the aggregated output of the clauses for each training example is central to our scheme. In a standard \ac{TM}, each training example $(X_i, y_i), 1 \le i \le q,$ is processed by first evaluating the clauses on $X_i$ and then obtaining the majority vote $v$ from Eq.~(\ref{eqn:aggregated_votes}). Here, $q$ is the total number of examples. The majority vote $v$ is then compared with the summation margin $T$ when $y=1$ and $-T$ when $y=0$, to produce the feedback to the \acp{TA} of each clause, explained in the previous section.

\begin{figure}[ht]
\vskip 0.2in
\begin{center}
\centerline{\includegraphics[width=\columnwidth]{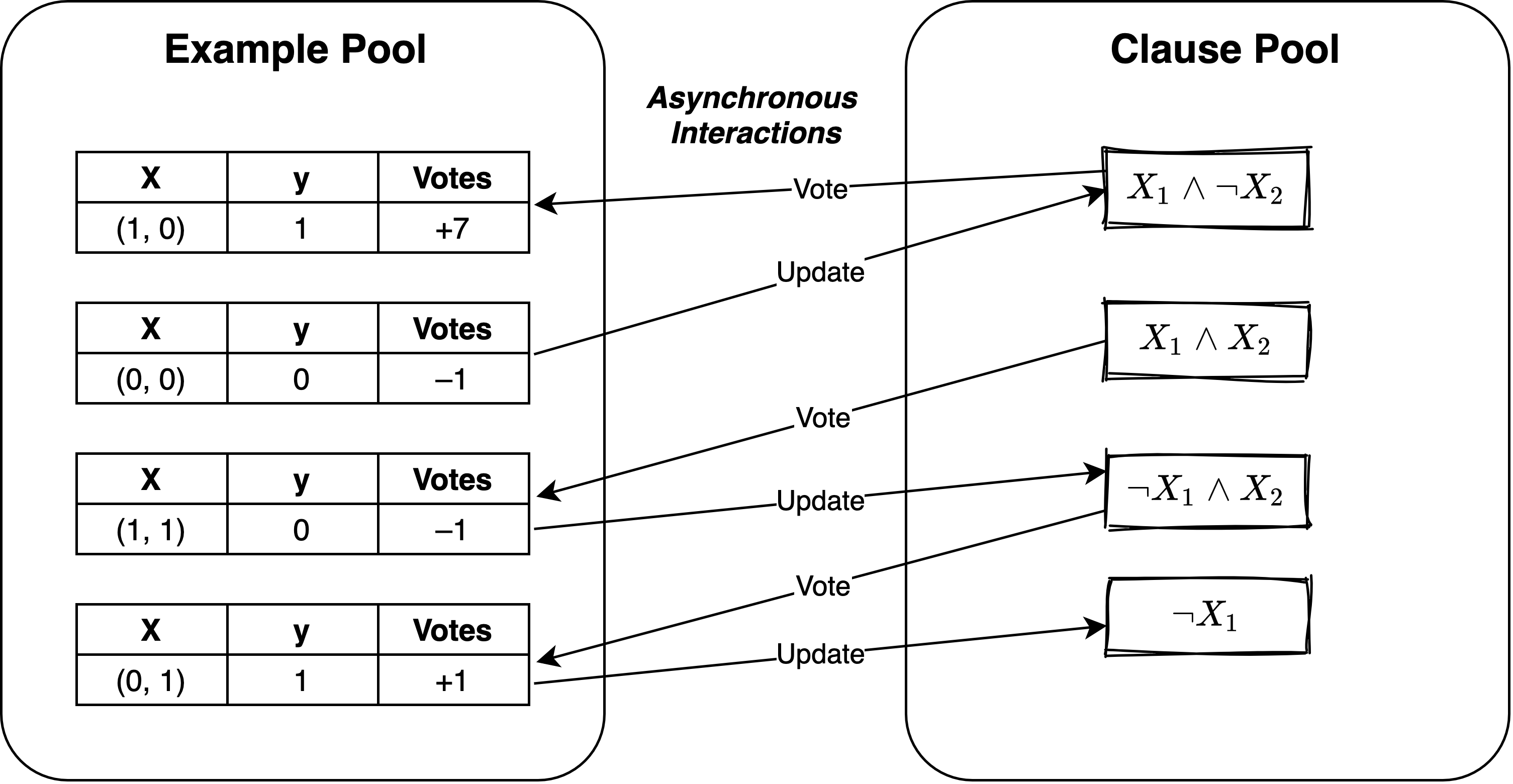}}
\caption{Parallel Tsetlin machine architecture.}
\label{figure:architecture_basic}
\end{center}
\vskip -0.2in
\end{figure}

As illustrated in Fig.~\ref{figure:architecture_basic}, to decouple the clauses, we now assume that the particular majority vote of example $X_i$ has been pre-calculated, meaning that each training example becomes a triple $(X_i, y_i, v_i)$, where $v_i$ is the pre-calculated majority vote. With $v_i$ in place, the calculation performed in Eq.~(\ref{eqn:aggregated_votes}) can be skipped, and we can go directly to give Type~I or Type~II feedback to any clause $C_j$, without considering the other clauses. This opens up for decentralized learning of the clauses, facilitating native  parallelization at all inference and learning steps. The drawback of this scheme, however, is that any time the composition of a clause changes after receiving feedback, all voting aggregates $v_i, 1 \le i \le q,$ become outdated. Accordingly, the standard learning scheme for updating clauses must be replaced.

\begin{algorithm}[tb]
%\scriptsize
\caption{Decentralized updating of clause}
\label{alg:decentralized_clause_updating}

\begin{algorithmic}
    \STATE {\bfseries Input: } Example pool $P$, clause $C_j$, positive polarity indicator \(p_j \in \{0, 1\}\), batch size \(b \in [1,\infty)\), voting margin \(T \in [1,\infty)\), pattern specificity \(s \in [1, \infty)\).
    \STATE {\bfseries Procedure: } UpdateClause : {$C_j, p_j, P, b, T, s$}.
    
    \FOR{$i=1$ {\bfseries to} $b$}
    \STATE $(X_i, y_i, v_i) \gets \mathrm{ObtainTrainingExample}(P)$
    \STATE $v_i^c \leftarrow \mathbf{clip}\left(v_i, -T, T\right)$
    \STATE $e = T - v_i^c \mathbf{~if~} y_i = 1 \mathbf{~else~} T + v_i^c$ 
    
    \IF{rand() $\le \frac{e}{2T}$}
    \IF{$y_i \mathbf{~xor~} p_j$}
    \STATE $C_j \gets$ TypeIIFeedback($X_i, C_j$)
    \ELSE
    \STATE $C_j \gets$ TypeIFeedback($X_i,  C_j, s$)
    \ENDIF 
        
    \STATE $o_{ij} \gets C_j(X_i)$
    \STATE $o_{ij}^{*} \gets \mathrm{ObtainPreviousClauseOutput}(i,j)$
    \IF{$o_{ij} \ne o_{ij}^{*}$}
    \STATE $\mathrm{AtomicAdd}(v_i, o_{ij} - o_{ij}^{*})$
    \STATE $\mathrm{StorePreviousClauseOutput}(i,j, o_{ij})$
    \ENDIF
    \ENDIF
    \ENDFOR

\end{algorithmic}

\end{algorithm}

\subsection{Decentralized Clause Learning}

Our decentralized learning scheme is captured by Algorithm~\ref{alg:decentralized_clause_updating}. As shown, each clause is trained independently of the other clauses. That is, each clause proceeds with training without taking other clauses into consideration. Algorithm~\ref{alg:decentralized_clause_updating} thus supports native parallelization because each clause now can run independently in its own thread.

\vspace{0.1in}
Notice further how the clause in focus first obtains a reference to the next training example $(X_i, y_i, v_i)$ to process, including the pre-recorded voting sum $v_i$ (Line 3). This example is retrieved from an example pool $P$, which is the storage of the training examples (centralized or decentralized).

The error of the pre-recorded voting sum $v_i$ is then calculated based on the voting margin $T$ (Line 5). The error, in turn, decides the probability of updating the clause. The  updating rule is the standard Type I and Type II \ac{TM} feedback, governed by the polarity $p_j$ of the clause and the specificity hyper-parameter $s$ (Lines 6-11).

The moment clause $C_j$ is updated, all recorded voting sums in the example pool $P$ are potentially outdated. This is because $C_j$ now captures a different pattern. Thus, to keep all of the voting sums $v_i$ in $P$ consistent with $C_j$, $C_j$ should ideally have been re-evaluated on all of the examples in $P$.

To partially remedy for outdated voting aggregates, the clause only updates the \emph{current} voting sum $v_i$. This happens when the calculated clause output $o_{ij}$ is different from the previously calculated clause output $o_{ij}^*$ (Lines 12-17). Note that the previously recorded output $o_{ij}^*$ is a single bit that is stored locally together with the clause. In this manner, the algorithm provides \emph{eventual consistency}. That is, if the clauses stop changing, all the voting sums eventually become correct.

\vspace{0.1in}
A point to note here is that there is no guarantee that the clause outputs in the parallel version will sum up to the same number as in the sequential version. The reason is that the updating of clauses is asynchronous, and the vote sums are not updated immediately when a clause changes. We use the term \emph{eventual consistency} to refer to the fact that if the clauses stop changing, eventually, the tallied voting sums become the exact sum of the clause outputs. Although not analytically proven, experimental results show that the two versions provide consistent final accuracy results after clause summation and thresholding.

Employing the above algorithm, the clauses access the training examples simultaneously, updating themselves and the local voting tallies in parallel. There is no synchronization among the clause threads, apart from atomic adds to the local voting tallies (Line 15). 
\section{Empirical Results}\label{sec:empirical}

In this section, we investigate how our new approach to \ac{TM} learning scales, including effects on training time and accuracy. We employ seven different data sets that represent diverse learning tasks, including regression, novelty detection, sentiment analysis, %reinforcement learning, 
semantic relation analysis, and word sense disambiguation. The data sets are of various sizes, spanning from $300$ to $100,000$ examples, $2$ to $20$ classes, and $6$ to $102,176$ features. We have striven to recreate \ac{TM} experiments reported by various researchers, including their hyper-parameter settings. For comparison of performance, we contrast with fast single-core \ac{TM} implementations\footnote{Retrieved from \href{https://github.com/cair/pyTsetlinMachine}{https://github.com/cair/pyTsetlinMachine}.} both with and without clause indexing \cite{gorji2020indexing}. Our proposed architecture is implemented in CUDA and runs on a Tesla V100 GPU (grid size $208$ and block size $128$). The standard implementations run on an Intel Xeon Platinum 8168 CPU at $2.70$ GHz. Code is available at \href{https://github.com/cair/ICML-Massively-Parallel-and-Asynchronous-Tsetlin-Machine-Architecture}{https://github.com/cair/ICML-Massively-Parallel-and-Asynchronous-Tsetlin-Machine-Architecture}.

We summarize the obtained performance metrics in Table~\ref{table_performance}. For better reproducibility, each experiment is repeated five times and the average accuracy and standard deviation are reported. We also report how much faster the CUDA TM executes compared with the indexed version.
\begin{table*}
\centering
\vskip 0.15in
\begin{center}
\begin{small}
\begin{sc}
\resizebox{.95\textwidth}{!}{
\begin{tabular}{l|l|l|l|l|l|l|c} 
\hline
\multicolumn{1}{c|}{ \textbf{Dataset} } & \multicolumn{2}{c|}{\textbf{TM indexed}} & \multicolumn{2}{c|}{\textbf{ TM non-indexed} } & \multicolumn{2}{c|}{\textbf{TM CUDA}} &\textbf{Speed up}\\ 
\cline{2-7}
\multicolumn{1}{c|}{} & \multicolumn{1}{c|}{Acc} & \multicolumn{1}{c|}{F1} & \multicolumn{1}{c|}{Acc} & \multicolumn{1}{c|}{F1} & \multicolumn{1}{c|}{Acc} & \multicolumn{1}{c|}{F1} &  \\ 
\hline
BBC Sports & 85.08 $\scriptstyle\pm$ 1.75  & 85.58 $\scriptstyle\pm$ 1.69  & 87.36 $\scriptstyle\pm$ 1.91  & 88.02 $\scriptstyle\pm$ 1.47  & 84.64 $\scriptstyle\pm$2.20  & 86.13 $\scriptstyle\pm$ 2.24  & 38.9$\times$ \\ 
\hline
20 Newsgroup & 79.37 $\scriptstyle\pm$ 0.25  & 80.38 $\scriptstyle\pm$ 0.92  & 82.33 $\scriptstyle\pm$ 0.28  & 82.89 $\scriptstyle\pm$ 0.34  & 79.00 $\scriptstyle\pm$ 0.46  & 78.93 $\scriptstyle\pm$ 0.44  & 49.3$\times$ \\ 
\hline
SEMEVAL & 91.9 $\scriptstyle\pm$ 0.16  & 75.29 $\scriptstyle\pm$ 0.25  & 92.51 $\scriptstyle\pm$ 0.03  & 77.48 $\scriptstyle\pm$ 1.46  & 92.02 $\scriptstyle\pm$ 0.54  & 76.27 $\scriptstyle\pm$ 0.53  &  1.7$\times$\\ 
\hline
IMDB & 88.42 $\scriptstyle\pm$ 2.05  & 88.39 $\scriptstyle\pm$ 2.16  & 88.2 $\scriptstyle\pm$ 3.14  & 88.13 $\scriptstyle\pm$ 3.44  & 89.92 $\scriptstyle\pm$ 0.23  & 88.90 $\scriptstyle\pm$ 0.24  & 34.6$\times$ \\ 
\hline
JAVA & 97.03 $\scriptstyle\pm$ 0.02  & 96.93 $\scriptstyle\pm$ 0.02  & 97.50 $\scriptstyle\pm$ 0.02  & 97.40 $\scriptstyle\pm$ 0.02  & 97.53 $\scriptstyle\pm$ 0.02  & 97.42 $\scriptstyle\pm$ 0.01  & 6.0$\times$ \\ 
\hline
Apple & 92.65 $\scriptstyle\pm$ 0.02  & 92.20 $\scriptstyle\pm$ 0.02  & 92.46 $\scriptstyle\pm$ 0.01  & 91.82 $\scriptstyle\pm$ 0.02  & 95.01 $\scriptstyle\pm$ 0.01  & 94.68 $\scriptstyle\pm$ 0.01  & 9.7$\times$ \\
\hline
\end{tabular}
}
\end{sc}
\end{small}
\end{center}
\vskip -0.1in
\caption{Performance on multiple data sets. Mean and standard deviation are calculated over 5 independent runs. Speed up is calculated as how many times faster is average execution time on CUDA implementation than on Indexed implementation. }
\label{table_performance}
\end{table*}

The focus in this section is to highlight the speed-ups achieved via parallelization of the TM architecture. However, for the sake of completeness, we also include corresponding results for SVM using ThunderSVM, which is an established GPU implementation of SVM \cite{wen2018thundersvm}. The performance results are compiled in Fig. \ref{table_thunderSVMtrainingtimecomparison}\footnote{Google's Colab is used to run the two CUDA models in the table for fair comparison.}. As seen, our TM parallelization is significantly faster than ThunderSVM for all the investigated datasets. Note that our parallelization solution leverages novel TM properties, allowing each TM clause to learn independently in its own thread. 
\begin{table*}[ht]
\begin{sc}
\small
\centering
\begin{tabular}{c|c|c|c|c|c|c|c|c}
\hline \textbf{CUDA Model} & \textbf{BBC Sports} & \textbf{20 NG} & \textbf{SEMEVAL} & \textbf{IMDB} & \textbf{JAVA(WSD)} & \textbf{Apple(WSD)} & \textbf{Regr1} & \textbf{Regr2}\\ \hline
ThunderSVM & 0.23s & 6.95s & 7.42s & 3.89s & 1.31s & 0.39s & 6.24s & 59.94s \\
TM & 0.13s & 3.61s & 4.27s & 3.36s &  0.74s & 0.28s & 0.52s & 4.92s\\

  \hline
\end{tabular}
\end{sc}
\caption{Comparison of training time between ThunderSVM and TM. The run time is an average of 5 experiments.}
\label{table_thunderSVMtrainingtimecomparison}
\end{table*}

\subsection{Regression}
We first investigate performance with \acp{RTM} using two datasets: Bike Sharing and BlogFeedback. 

The Bike Sharing dataset contains a 2-year usage log of the bike sharing system Captial Bike Sharing (CBS) at Washington, D.C., USA. The total number of rental bikes (casual and registered), is predicted based on other relevant features such as time of day, day of week, season, and weather. In total, $16$ independent features are employed to predict the number of rental bikes. Overall, the dataset contains $17,389$ examples. More details of the Bike Sharing dataset can be found in \cite{fanaee2014event}.

The BlogFeedback dataset considers predicting the number of blog posts comments for the upcoming $24$ hours. The data has been gathered from January, 2010 to March, 2012. In total, BlogFeedback contains $60,021$ data samples, each with $281$ features. In this study, we employ the first $20,000$ data samples. For both of the datasets, we use $80\%$ of the samples for training and the rest for evaluation.

\begin{figure}[!t]

\begin{center}
\centerline{\includegraphics[width=0.8\columnwidth]{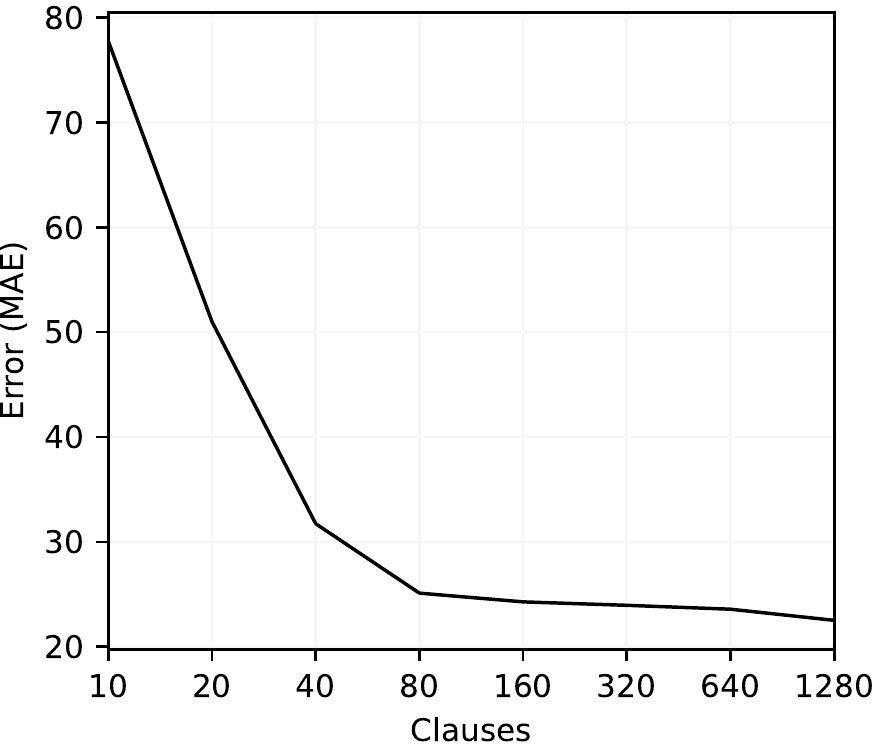}}
\caption{\label{reg0} MAE vs. \#clauses on Bike Sharing.}
\end{center}

\end{figure}

We first study the impact of the number of clauses on prediction error, measured by Mean Absolute Error (MAE). As illustrated in Fig.~\ref{reg0} for Bike Sharing, increasing the number of clauses (\#clauses in x-axis) decreases the error by allowing the \ac{RTM} to capture more detailed sub-patterns.

\begin{figure}[!t]
%\vskip 0.2in
\begin{center}
\centerline{\includegraphics[width=0.85\columnwidth]{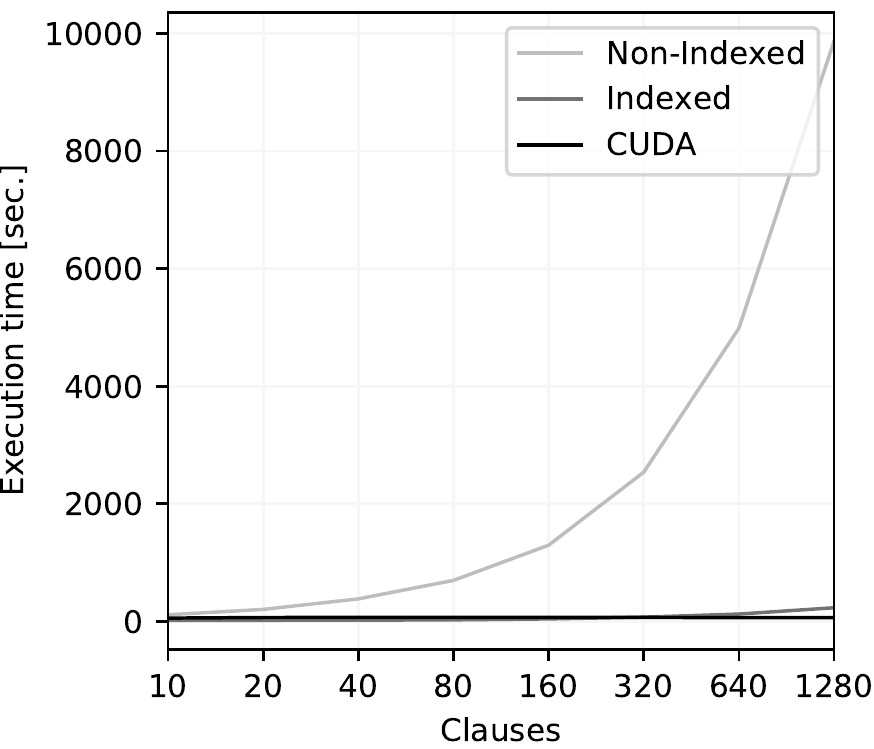}}
\caption{\label{reg3} Execution time vs. \#clauses on Bike Sharing.}
\end{center}
\vskip -0.3in
\end{figure}

A larger number of clauses results in increased computation time. Fig.~\ref{reg3} captures how execution time increases with the number of clauses for the three different implementations. For the non-indexed \ac{RTM}, each doubling of the number of clauses also doubles execution time. However,  the indexed \ac{RTM} is less affected, and is slightly faster than the CUDA implementation when less than $300$ clauses are utilized. With more than 300 clauses, the CUDA implementation is superior, with no significant increase in execution time as the number of clauses increases. For instance, using $1280$ clauses, the CUDA implementation is roughly $3.64$ times faster than the indexed version. This can be explained by the large number of threads available to the GPU and the asynchronous operation of the new architecture.

%-------------------------------------------------------------------------

%-------------------------------------------------------------------------

\begin{figure}[!t]
%\vskip 0.2in
\begin{center}
\centerline{\includegraphics[width=0.8\columnwidth]{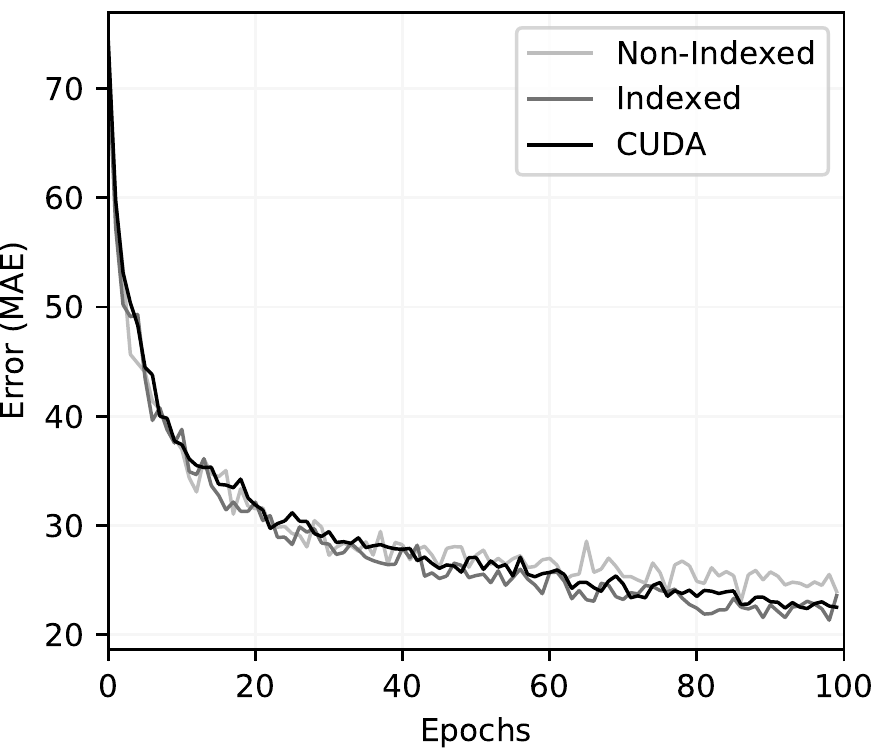}}
\caption{\label{reg1} MAE over epochs on Bike Sharing.}
\end{center}
\vskip -0.2in
\end{figure}

\begin{figure}[!t]
%\vskip 0.2in
\begin{center}
\centerline{\includegraphics[width=0.8\columnwidth]{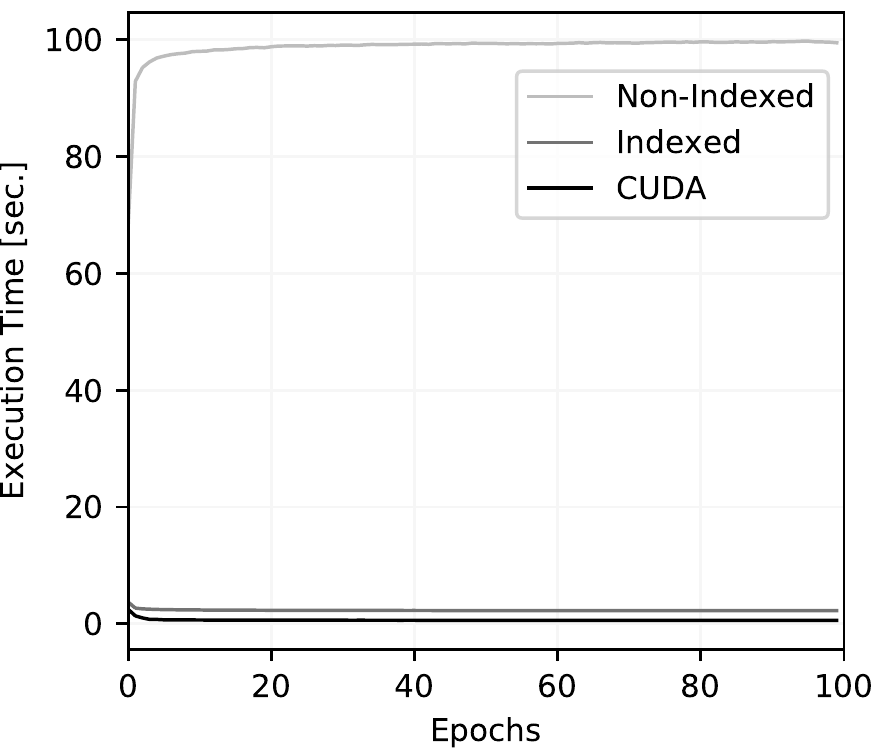}}
\caption{\label{reg2} Per Epoch Execution time over epochs on Bike Sharing.}
\end{center}
\vskip -0.3in
\end{figure}

Looking at how MAE and execution time vary over the training epochs for Bike Sharing (Fig.~\ref{reg1} and Fig.~\ref{reg2}, respectively), we observe that MAE falls systematically across the epochs, while the execution time remains stable (employing $T=1280$, $s=1.5$, $n=1280$). Execution on BlogFeedback exhibits similar behavior. Finally, the mean value of the MAEs for each method are similar (Table~\ref{table_regressionperformance}) across 5 independent runs, indicating no significant difference in learning accuracy.

%The execution time of the CUDA implementation can be further controlled by modifying the number of threads, e.g., by changing the block size. Fig.~\ref{reg4} shows the variation of execution time with increase of block size for Bike Sharing: increasing block size reduces execution time for $5,000$ clauses, while having limited effect on $1,280$. This means that more clauses are needed to leverage the increase in number of threads.

\begin{table*}
\vskip 0.15in
\setlength{\tabcolsep}{0.1em}
\begin{center}
\begin{small}
\begin{sc}
\begin{tabular}{l|c|c|c|c} 
\hline
\multicolumn{1}{c|}{ \textbf{Datasets} } & \begin{tabular}[c]{@{}c@{}}\textbf{TM}\\\textbf{indexed } \end{tabular} & \begin{tabular}[c]{@{}c@{}}\textbf{TM}\\\textbf{non-indexed} \end{tabular} & \begin{tabular}[c]{@{}c@{}}\textbf{TM}\\\textbf{CUDA} \end{tabular} &  \textbf{Speed up}  \\ 
\hline
Bike Sharing & 23.5$\scriptstyle\pm$0.04  & 22.5$\scriptstyle\pm$0.12 & 23.9$\scriptstyle\pm$0.08 & 156.4$\times$ \\ 
\hline
BlogFeedback. & 3.91$\scriptstyle\pm$0.00 & 3.74$\scriptstyle\pm$0.03 & 3.88$\scriptstyle\pm$0.02 & 211.6$\times$ \\
\hline
\end{tabular}
\end{sc}
\end{small}
\end{center}
\vskip -0.1in
\caption{MAE with confidence interval, and Speed up on two regression datasets, calculated over 5 independent runs}
\label{table_regressionperformance}
\end{table*}

%still waiting for the BlogFeedback results.
%-------------------------------------------------------------------------

%-------------------------------------------------------------------------

%-------------------------------------------------------------------------

%\begin{figure}[h]
%\vskip 0.2in
%\begin{center}
%\centerline{\includegraphics[width=0.8\columnwidth]{graphs_darshana/Energy_blocksVStime.pdf}}
%\caption{\label{reg4} CUDA Exec. time vs. block size on Bike Sharing with differing \#clauses.}
%\end{center}
%\vskip -0.3in
%\end{figure}

Comparing MAE with comparable contemporary methods on Bike Sharing (Regr1), we observe that the TM is competitive -- TM: 23.9; ANN: 27.4; and  SVR: 22.6. Similarly, for BlockFeedback (Regr2) we have -- TM: 3.88; ANN: 7.2; and SVR: 6.0.

\subsection{Novelty Detection}
Novelty detection is another important machine learning task. Most supervised classification approaches assume a closed world, counting on all classes being present in the data at training time. This assumption can lead to unpredictable behaviour during operation, whenever novel, previously unseen, classes appear. We here investigate \ac{TM}-based novelty detection, as proposed by \citet{bhattarai2021novelty, bhattarai2021wordlevel}. To this end, we use the two datasets: 20 Newsgroup and BBC Sports. In brief, we use the class voting sums (Section \ref{sec:basics}) as features measuring novelty. We then employ 
a Multilayer perceptron~(MLP) for novelty detection that adopts the class voting sums as input.

The BBC sports dataset contains $737$ documents from the BBC sport website, organized in five sports article categories and collected from 2004 to 2005. Overall, the dataset encompasses a total of $4,613$ terms.  For novelty classification, we designate the classes ``Cricket" and ``Football" as known and ``Rugby" as novel. We train on the known classes, which runs for $100$ epochs with $5,000$ clauses, margin $T$ of $100$, and specificity $s$ of $15.0$. The training times for both indexed and non-indexed \acp{TM} are high compared with that of CUDA TM, which is around $39$ times faster.
The 20 Newsgroup dataset contains $18,828$ documents with $20$ classes. The classes ``comp.graphics" and ``talk.politics.guns" are designated as known, and ``rec.sport.baseball" is considered novel. We train the \ac{TM} for $100$ epochs with a margin $T$ of $500$, $10,000$ clauses and specificity $s=25.0$. The CUDA \ac{TM} implementation is here about $49$ times faster than the other versions.

\begin{figure}[!t]
%\vskip 0.2in
\begin{center}
\centerline{\includegraphics[width=0.8\columnwidth]{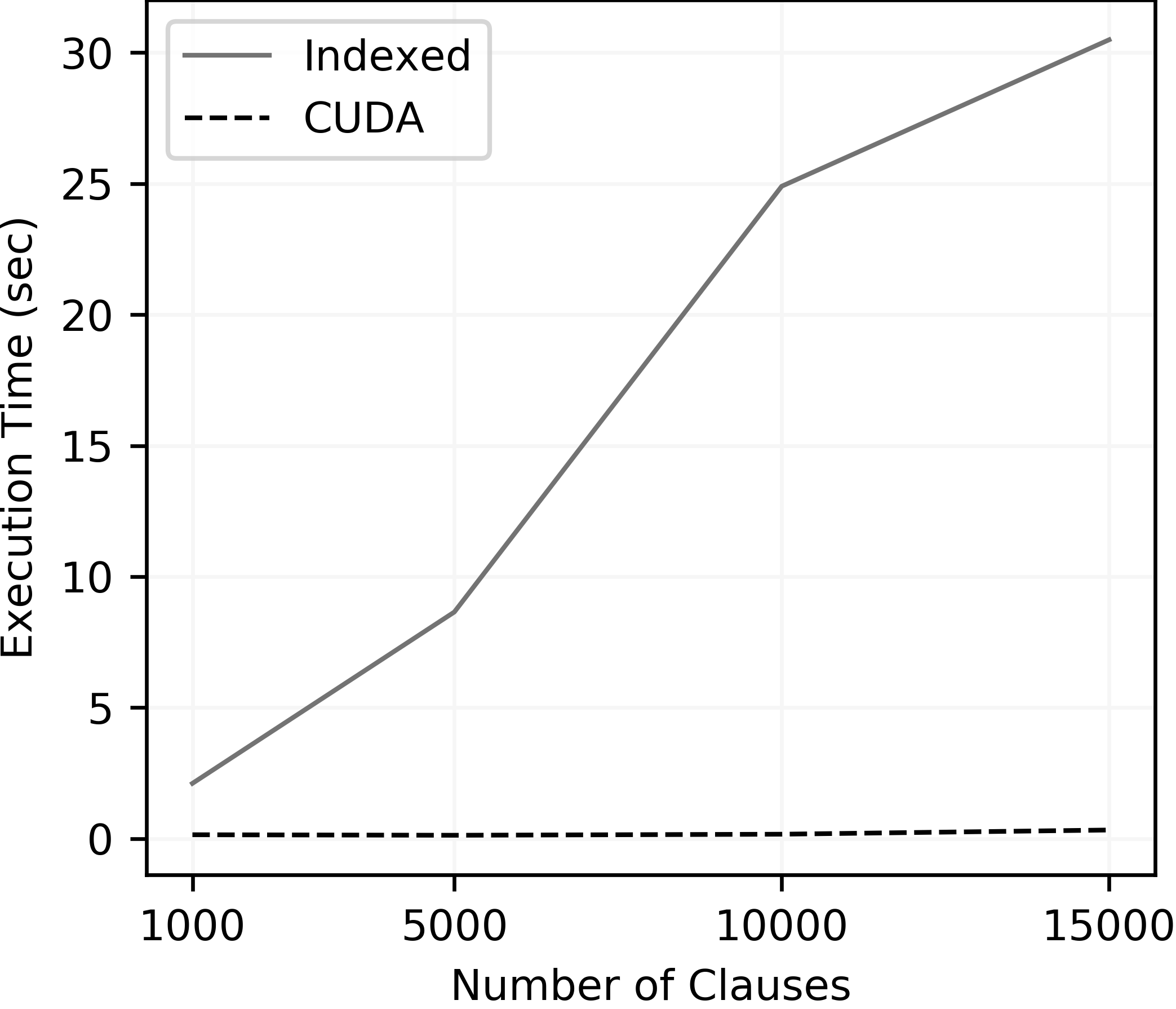}}
\caption{Execution time vs. \#clauses on BBC Sports.}
\label{fig_bbc_clauses_time}
\end{center}
\vskip -0.3in
\end{figure}

To assess scalability, we record the execution time of both the indexed and the CUDA \ac{TM} while increasing the number of clauses (Fig.~\ref{fig_bbc_clauses_time}). For the indexed \ac{TM}, the execution time increases almost proportionally with the number of clauses, but no such effect is noticeable for the CUDA \ac{TM}.

The novelty scores generated by the \ac{TM} are passed to a Multilayer Perceptron with hidden layer sizes ($100$, $30$) and RELU activation functions, trained using stochastic gradient descent. As shown in Table~\ref{table_performance}, for both datasets, the non-indexed \ac{TM} slightly outperforms the other \ac{TM} versions, while the indexed and CUDA \acp{TM} have similar accuracy. These differences can be explained by the random variation of \ac{TM} learning (i.e., the high standard deviations reported in Table~\ref{table_performance}).

\vspace{0.1in}
\citet{bhattarai2021novelty} investigate the competitiveness of TMs on novelty detection. For BBC Sports they report the following accuracies -- TM: $89.47\%$; One-class SVM: $83.53\%$; and Average KNN: $55.54\%$. Likewise, for 20Newsgroup they report -- TM: $82.51\%$; One-class SVM: $83.70\%$; and Average KNN: $81\%$.

\subsection{Sentiment and Semantic Analysis}
We adopt the \emph{SemEval 2010 Semantic Relations} \cite{hendrickx2009semeval} and the \emph{ACL Internet Movie Database (IMDb)} \cite{maas2011learning} datasets to explore the performance of the \ac{TM} implementations for a large number of sparse features, as proposed by \citet{saha2020causal}.

The  SEMEVAL dataset focuses on identifying semantic relations in text. The dataset has $10,717$ examples, and we consider each to be annotated to contain either the relation Cause-Effect or not. The presence of an unambiguous causal connective is indicative of a sentence being a causal sentence \cite{xuelan1992expressing}. For each \ac{TM}, we use $40$ clauses per class to identify this characteristic of causal texts. The IMDb dataset contains $50,000$ highly polar movie reviews, which are either positive or negative. Due to the large variety and combination of possibly distinguishing features, we assign $7,000$ clauses to each class. For both datasets we use unigrams and bigrams as features.

\begin{figure}[!t]
%\vskip 0.2in
\begin{center}
\centerline{\includegraphics[width=0.8\columnwidth]{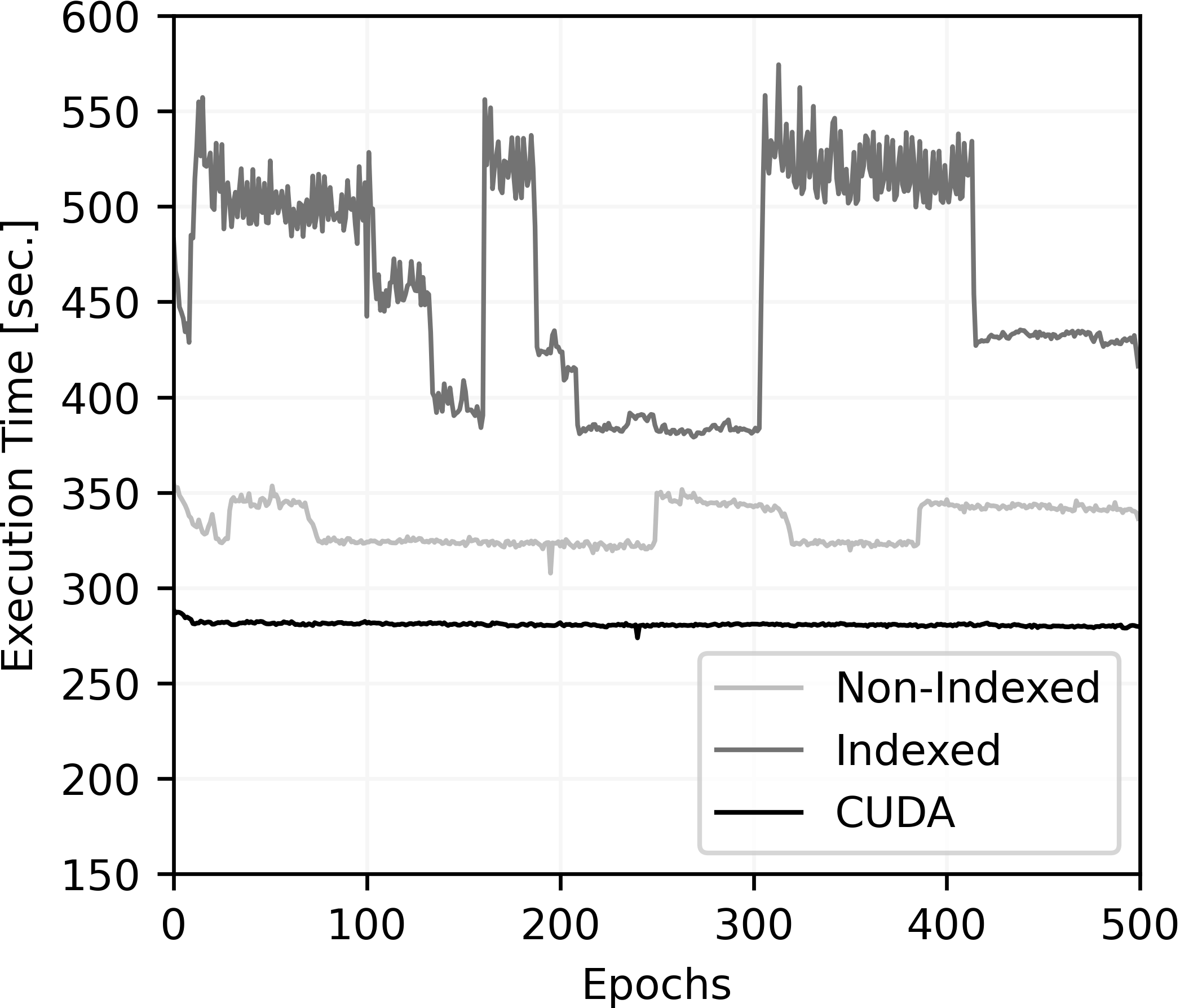}}
\caption{\label{fig_traintime_semeval} Per Epoch Execution time over epochs on SEMEVAL.}
\end{center}
\vskip -0.3in
\end{figure}

\begin{figure}[!t]
\vskip 0.2in
\begin{center}
\centerline{\includegraphics[width=0.8\columnwidth]{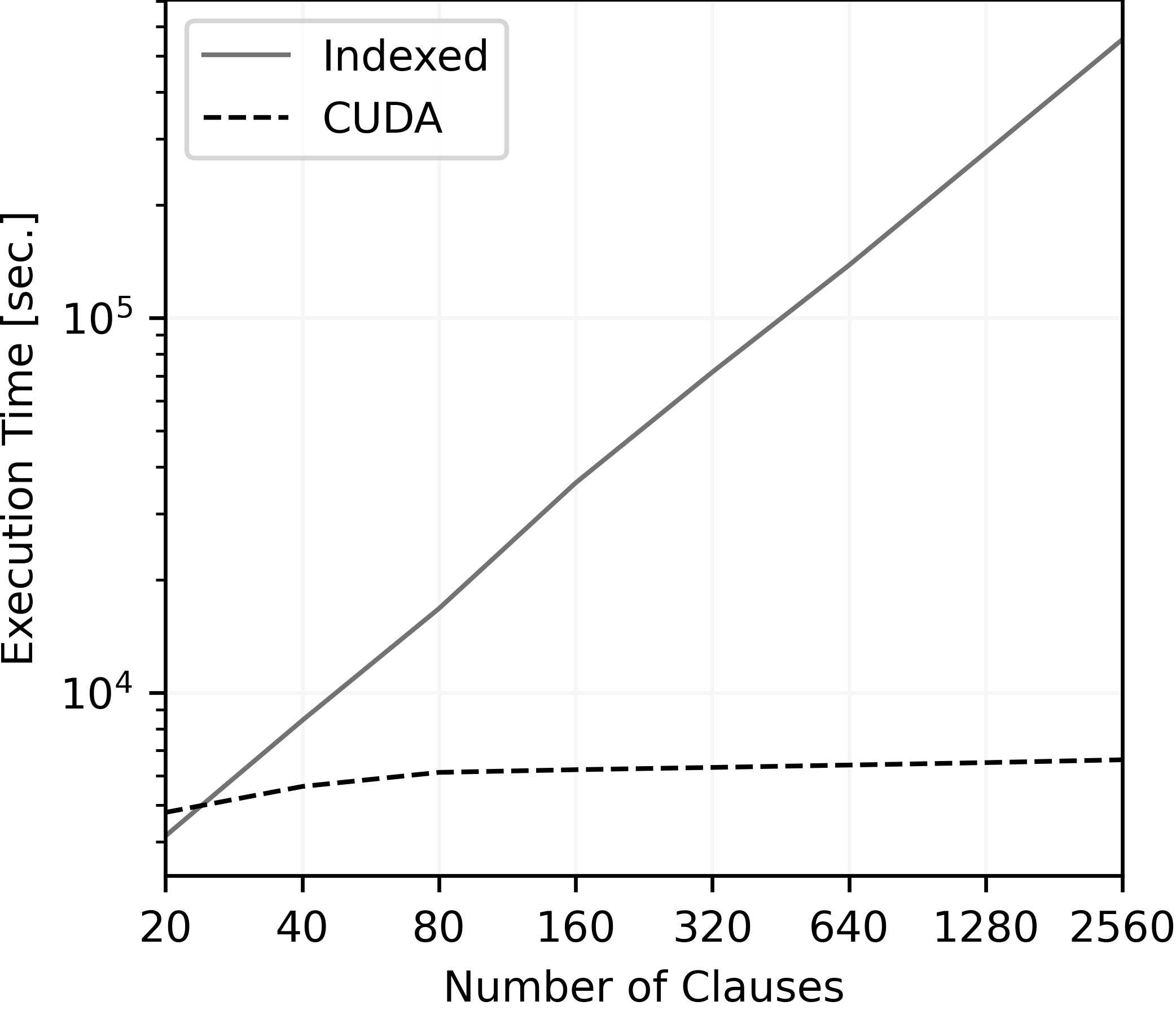}}
\caption{Execution time vs. \#clauses on SEMEVAL.}
\label{figtraintime_clauses_semeval}
\end{center}
\vskip -0.3in
\end{figure}

As noted in Table \ref{table_performance}, the accuracy obtained by the CPU (non-indexed) and the CUDA implementations are comparable on the SEMEVAL dataset, while the indexed \ac{TM} performs slightly poorer. However, the execution time is much lower for the CUDA version than the other two (Fig.~\ref{fig_traintime_semeval}). This is further shown in Fig.~\ref{figtraintime_clauses_semeval}. Clearly, the CPU-based \ac{TM} with indexing takes an increasing amount of time to execute as the number of clauses grows, but no such effect is observed with CUDA \ac{TM}. More specifically,  the execution time increases only by $40\%$ when the number of clauses goes from $20$ to $2,560$.

With the IMDB dataset, the CUDA version performs better in terms of accuracy, with less variance compared to the CPU versions (Table \ref{table_performance}). It exhibits similar behaviour as in the SEMEVAL dataset with respect to execution time over increasing number of epochs. 
%From approximately $7,000$ clauses and onwards, however, we observe proportionally increasing execution time, e.g., the execution time doubles going from $7,000$ to $14,000$ clauses (Fig.~\ref{figtraintime_clauses_imdb}). This behavior may be caused by the Tesla V100 GPU that has $5,120$ cores. 

%We also show how the change in CUDA block size affects the execution time, given a particular number of clauses in Fig.~\ref{fig_blocksize_semeval}. With less number of clauses, there are no benefits to use a larger block size. When a large number of clauses are employed, a larger block size effectively parallelizes the workload of the \ac{TM}, reducing the execution time.
%%%%%%%%%%%%%%%%%%%%%%%%%%%%%%%%%%%%%%
%%SEMEVAL
%%%%%%%%%%%%%%%%%%%%%%%%%%%%%%%%%%%%%%
%\subsubsection{SEMEVAL}

%%%%%%%%%%%%%%%%%%%%%%%%%%%%%%%%%%%%%
%% 1: Accuracy Indexed vs. Non-Indexed vs.  CUDA
%%%%%%%%%%%%%%%%%%%%%%%%%%%%%%%%%%%%%%
%\begin{figure}[t]
%\centering
%{\includegraphics[scale=0.7]
%{graphs_rupsa/semeval_accuracy.png}}
%\caption{\label{fig_acc_semeval} Comparison of testing accuracy on %SEMEVAL dataset for Non-indexed, Indexed and CUDA Tsetlin machine.}
%\end{figure}

%%%%%%%%%%%%%%%%%%%%%%%%%%%%%%%%%%%%%%
%% 2: Execution Times Indexed vs. Non-Indexed vs.  CUDA
%%%%%%%%%%%%%%%%%%%%%%%%%%%%%%%%%%%%%%

%%%%%%%%%%%%%%%%%%%%%%%%%%%%%%%%%%%%%%
%% 3: Execution Times with Incr Clauses Indexed vs.  CUDA
%%%%%%%%%%%%%%%%%%%%%%%%%%%%%%%%%%%%%%

%%%%%%%%%%%%%%%%%%%%%%%%%%%%%%%%%%%%%%
%% 3: Execution Times with Incr Clauses Indexed vs.  CUDA
%%%%%%%%%%%%%%%%%%%%%%%%%%%%%%%%%%%%%%

%\begin{figure}[ht]
%%\vskip 0.2in
%\begin{center}
%\centerline{\includegraphics[width=0.8\columnwidth]{graphs_rupsa/increaseclauses_imdb_training_times_log.png}}
%\caption{Execution time vs. \#clauses on IMDb.}
%\label{figtraintime_clauses_imdb}
%\end{center}
%\vskip -0.2in
%\end{figure}

%%%%%%%%%%%%%%%%%%%%%%%%%%%%%%%%%%%%%%
%% 4: Execution Time CUDA blocksize
%%%%%%%%%%%%%%%%%%%%%%%%%%%%%%%%%%%%%%

%\begin{figure}[ht]
%%\vskip 0.2in
%\begin{center}
%\centerline{\includegraphics[width=0.8\columnwidth]{graphs_rupsa/blocksize_trainingtime_semeval_log.png}}
%\caption{CUDA TM Execution time vs. block size on SEMEVAL with differing \#clauses.}
%\label{fig_blocksize_semeval}
%\end{center}
%\vskip -0.2in
%\end{figure}
%%%%%%%%%%%%%%%%%%%%%%%%%%%%%%%%%%IMDB%%%%%%%%%%%%%%%%%%%
%%%%
%%%%
%%%%%%%%%%%%%%%%%%%%%%%%%%%%%%%%%%%%%%%%%%%%%%%%%%%%%%%%%
%\subsubsection{IMDB}

%%%%%%%%%%%%%%%%%%%%%%%%%%%%%%%%%%%%%%
%% 1: Accuracy Indexed vs. Non-Indexed vs.  CUDA
%%%%%%%%%%%%%%%%%%%%%%%%%%%%%%%%%%%%%%
%\begin{figure}[t]
%\centering
%{\includegraphics[scale=0.7]
%{graphs_rupsa/imdb_accuracy.png}}
%\caption{\label{fig_acc_imdb} Comparison of testing accuracy on IMDb dataset for Non-indexed, %Indexed and CUDA Tsetlin machine.}
%\end{figure}

The following provides a comparison of accuracy, compiled from related literature. For Semantic Relation Classification (SemEval) TM accuracy is again competitive (TM: 92.6\%; RandomForest: 88.76\%; NaiveBayes: 87.57\%; SVM: 92.3\%; and CNN-LSTM: 90.73\%).
The same goes for IMDB Sentiment Analysis (TM: 90.5\%; NBOW: 83.62\%; CNN-BiLSTM: 89.54\%; Tree-LSTM: 90.1\%;  Self-AT-LSTM: 90.34\%; and DistilBERT: 92.82\%).

\subsection{Word Sense Disambiguation}

%%%%%%%%%%%%%%%%%%%%%%%%%%%%%%%%%%%%%%
%% 3: Execution Times with Incr Clauses Indexed vs.  CUDA
%%%%%%%%%%%%%%%%%%%%%%%%%%%%%%%%%%%%%%

%\begin{figure}[ht]
%%\vskip 0.2in
%\begin{center}
%\centerline{\includegraphics[width=0.8\columnwidth]{graphs_rohan/apple_clause_time.png}}
%\caption{\label{apple_time} Comparison of execution time on Apple dataset for Non-indexed, Indexed and CUDA TM.}
%\label{apple_clause_time}
%\end{center}
%\vskip -0.2in
%\end{figure}

Word Sense Disambiguation (WSD) is a vital task in NLP~\cite{Navigli2009WordSD} that consists of distinguishing the meaning of homonyms -- identically spelled words whose sense depends on the surrounding context words. We here perform a quantitative evaluation of the three \ac{TM} implementations using a recent WSD evaluation framework~\cite{loureiro2020language} based on WordNet. We use a balanced dataset for coarse-grained classification, focusing on two specific domains. The first dataset concerns the meaning of the word ``Apple'', which here has two senses: ``apple\_inc.'' (company) and ``apple\_apple'' (fruit). The other dataset covers the word ``JAVA'', which has the two senses: ``java\_java'' (geographical location) and ``java\_comp.'' (computer language). The Apple dataset has $2,984$ samples split into training and testing samples of $1,784$ and $1,200$, respectively. The JAVA dataset has $5,655$ samples split into $3,726$ and $1,929$ samples, for training and testing. For preprocessing, we filter the stop words and stem the words using the Porter Stemmer to reduce the effect of spelling mistakes or non-important variations of the same word. To build a vocabulary (the feature space), we select the $3,000$ most frequent terms. The number of clauses $n$, margin $T$, and specificity $s$ used are $300$, $50$, $5.0$ respectively, for both datasets.

The accuracy and F1 scores of the non-indexed and indexed \acp{TM} are quite similar for the Apple dataset (Table~\ref{table_performance}). However, the CUDA TM outperforms both of them by a significant margin. In the case of JAVA dataset, the performance is comparable for all three, and CUDA \ac{TM} is slightly better. The reader is requested to refer to \cite{yadav2021wordsense} for further details on \ac{TM}-based WSD.

%On the other hand, from figure \ref{apple_time} and \ref{java_time}, we see that the execution time for non-indexed and indexed TM is quite similar, while CUDA TM exhibits a significantly less execution time. 
%Again, we observe no significant increase in execution time with respect to increasing number clauses for the CUDA \ac{TM}. The indexed \ac{TM}, on the other hand, experiences a substantial increase in computation time (Fig.~\ref{apple_clause_time}).

The following is a comparison of accuracy with respect to related literature \cite{loureiro2020language}. For Apple we have TM: $95.1\%$; ThunderSVM: $80.32\%$; FastTextBase-1NN: $96.3\%$; FastTextCommonCrawl-1NN: $97.8\%$; and BERTbase: $99.0\%$. Correspondingly, for JAVA we get TM: $97.53\%$; ThunderSVM: $93.62\%$; FastTextBase-1NN: $98.7\%$; FastTextCommonCrawl-1NN: $99.5\%$; and  BERTbase: $99.6\%$. Note that the latter TM results are from our current paper. Another recent paper, however, shows that better hyperparameters can make the \ac{TM} outperform FastTextBase-1NN, with the TM reaching $97.58\%$ and $99.38\%$ for Apple and JAVA, respectively \cite{yadav2021wordsense}.

\subsection{Analysis of Experiment Results}
The above results support our initial claims that the proposed \ac{TM} architecture provides learning speeds that are nearly constant for a reasonable number of clauses. Beyond a sufficiently large clause number, the computation time increases at approximately the same rate as the increase in number of clauses. The number of clauses to use highly depends on the dataset. In our case, using up to $7,000$ clauses provides high accuracy for all the datasets, simultaneously demonstrating almost constant training time overall. A further increase does not significantly improve accuracy. I.e., the number of clauses was decided empirically to provide a general benchmark. Employing $7,000$ clauses also allows us evaluate the capability of leveraging the $5,120$ cores on the GPU. Hence, considering both datasets and available cores, $7,000$ clauses are enough to demonstrate the almost constant time scaling, until exhausting the cores. However, we also extended the clauses to $15,000$ as shown in Fig. 7, and the execution time remained almost the same, thereby validating excellent exploitation of the available cores. To achieve this, our scheme adds $1$ bit per example per clause for tallying, enabling desynchronization.  TMs are further reported to use less memory than ANNs \cite{lei2020arithmetic}.
\section{Conclusions and Future Work}\label{sec:conclusion}

In this paper, we proposed a new approach to \ac{TM} learning, to open up for massively parallel processing. Rather than processing training examples one-by-one as in the original \ac{TM}, the clauses access the training examples simultaneously, updating themselves and local voting tallies in parallel. The local voting tallies allow us to detach the processing of each clause from the rest of the clauses, supporting decentralized learning. There is no synchronization among the clause threads, apart from atomic adds to the local voting tallies. Operating asynchronously, each team of \acp{TA} most of the time executes on partially calculated or outdated voting tallies.

%The main conclusions of the paper can be summarized as follows: 
%\begin{itemize}
%\item Our decentralized \ac{TM} architecture copes remarkably with working on outdated data, resulting in no significant loss in learning accuracy across diverse learning tasks (regression, novelty detection,
%%reinforcement learning,
%semantic relation analysis, and word sense disambiguation).
%\item Learning time is almost \emph{constant} for reasonable clause amounts (employing from $20$ to $7~000$ clauses on a Tesla V100 GPU).
%\item For sufficiently large clause numbers, computation time increases approximately proportionally.
%\end{itemize}
%Our parallel and asynchronous architecture thus allows processing of more massive data sets and operating with more clauses for higher accuracy, significantly increasing the impact of logic-based machine learning.

Based on the numerical results, our main conclusion is that \ac{TM} learning is very robust towards relatively severe distortions of communication and coordination among the clauses. Our results are thus compatible with the findings of \citet{shafik2020explainability}, who shows that \ac{TM} learning is inherently fault tolerant, completely masking stuck-at faults. 

In our future work, we will investigate the robustness of \ac{TM} learning further, which includes developing mechanisms for heterogeneous architectures and more loosely coupled systems, such as grid-computing.

\bibliography{Reference}
\bibliographystyle{icml2021}

\end{document}